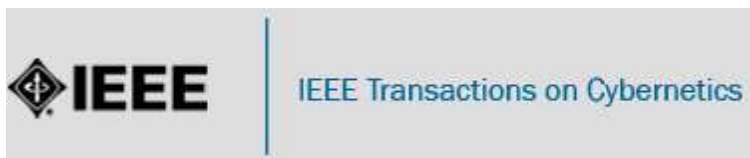

# On robot compliance. A cerebellar control approach

SCHOLARONE™
Manuscripts

# On robot compliance. A cerebellar control approach

Ignacio Abadía, Francisco Naveros, Jesús A. Garrido, Eduardo Ros, Niceto R. Luque

*Abstract*—**The work presented here is a novel biological approach for the compliant control of a robotic arm in real time (RT). We integrate a spiking cerebellar network at the core of a feedback control loop performing torque driven control. The spiking cerebellar controller provides torque commands allowing for accurate and coordinated arm movements. To compute these output motor commands, the spiking cerebellar controller receives the robot's sensorial signals, the robot's goal behaviour, and an instructive signal. These input signals are translated into a set of evolving spiking patterns, representing univocally a specific system state at every point of time. Spike Timing-Dependent Plasticity (STDP) is then supported, allowing for building adaptive control. The spiking cerebellar controller continuously adapts the torque commands provided to the robot from experience as STDP is deployed. Adaptive torque commands, in turn, help the spiking cerebellar controller to cope with built-in elastic elements within the robot's actuators mimicking human muscles (inherently elastic). We propose a natural integration of a bio-inspired control scheme, based on the cerebellum, with a compliant robot. We prove that our compliant approach outperforms the accuracy of the default factory-installed position control in a set of tasks used for addressing cerebellar motor behaviour: controlling six degrees of freedom (DoF) in (i) smooth movements, (ii) fast ballistic movements and (iii) unstructured scenario compliant movements.**

## I. INTRODUCTION

As far back as the 17th century, John Donne's famous "No man is an island, entire of itself" contrasted the idea of a self-sufficient object (an island) that does not interact with other objects against the importance of the interactions amongst people and their outcomes for society. Human being's ability to interact with others has facilitated their collaboration and coordination of their actions towards achieving common goals. For many daily tasks, collaboration between humans also involves physical human interaction, a technical term for a physical communication between two or more individuals in a shared context. The emergence of humanoid robots by the mid-nineties brought a new "individual" to interact within this shared context, thus extending the human-to-human interaction theory to humanrobot interaction theory, i.e. Human-Robot Interaction (HRI).

Efforts to HRI are being incrementally devoted over the last years[1], addressing new application domains in which new generations of robots begin to coexist and physically interact with humans (e.g., rehabilitation therapy [2], social interaction [3], education [4]) in contrast to the traditional well-structured industrial robotic scenarios lacking HRI. Physical HRI implies robots operating in complex unstructured environments in which human actions cannot be modelled; thus, demanding robot behaviour to be autonomous, reactive under unpredicted actions, adaptive and safe (i.e. human-like behaviour) [5]. The achievement of such compliant behaviour can be addressed considering different design aspects of robotic hardware (rigid vs. flexible materials, elastic actuators, low power actuators, etc.) and software (position vs. torque control, adaptive control systems, etc.).

Regarding hardware design, robots can be equipped with passive intrinsic compliance by means of different elastic components, muscle like actuators and/or soft materials. This approach, taking biology as an inspiration, offers a compliant alternative to classical rigid-bodied robots. Yet, traditional position control methods are not of direct application in the presence of elastic materials whose mathematical modelling is almost intractable, thus demanding new control strategies [6, 7]. These traditional methods offer excellent accuracy for industrial rigid-bodied robots in well-structured environments (e.g. automated car factories) where HRI is explicitly avoided since neither safety nor compliance can be guaranteed. Compliance demands torque control, and torque control strategies based on dynamics modelling cannot be efficiently applied since the nonlinearities of elastic components make detailed modelling extremely complex [8]. Finding a solution for controlling biologically inspired robots carrying elastic components and low power actuators shall directly benefit from a better understanding of biological motor control itself.

The control mechanisms encountered in biology are involved in a continuous learning process to cope with the complexity and changes in the body structure and dynamics. Artificial Intelligence (AI) can be used to replicate this learning process; in particular, widely used Artificial Neural Networks (ANNs) have been proposed and tested as a solution

The work of I. Abadía was supported by U. Granada and Junta Andalucía-FEDER. The work of F. Naveros and E. Ros were supported by the Spanish National Grant (MINECO-FEDER TIN2016-81041-R) and the EU HBP-SGA2, (H2020-RIA, 785907). The work of N. R. Luque was supported by a Juan de la Cierva Spanish fellowship (IJCI-2016-27385). The work of J. A. Garrido was supported by U. Granada under a Young Researchers fellowship. (*I. Abadía and F. Naveros contributed equally to this work*).

I. Abadia, F. Naveros, N. R. Luque, J. A. Garrido and E. Ros are with the Research Centre for Information and Communication Technologies, Dept. Computer Architecture and Technology, U. Granada, 18014 Granada, Spain (e-mail: iabadia, fnaveros, jesusgarrido, eros, nluque at ugr.es).

for the control of these compliant robots without requiring prior knowledge of the robot dynamics [8, 9]. ANNs are vaguely inspired in the functioning of their biological neural network counterparts. They consist of interconnected computational units, called artificial neurons, whose entry information travels from one computational unit to another across the ANN. The entry information is processed, at a neuron level, via some non-linear function of the sum of neuron inputs and then it is transmitted through the neuron connections, i.e. typically represented by a real number. Neuron connections are adjusted as learning proceeds. ANNs are designed to address problems by considering well-structured data typically using standard analogue representations for neural activity. They lack the ability to serve as the linkage between biological neural coding and movement coordination, thus side-lining any attempt at drawing biological analogies. Spiking Neural Networks (SNNs), also called the third generation of neural networks, constitute a more biologically plausible approach of neural networks as they model the information transfer and processing as occurs in biological neurons, i.e. via the precise timing of spikes (discrete events at points in time) [10]. Torque control deals with the robot inner dynamics, that is, the evolution through time of a physical system. This makes SNNs use of temporal coding adequate for capturing the temporal evolution of analogue sensorimotor signals [11], a pivotal feature in motor control and movement coordination [12]. SNNs intrinsic characteristics make them a suitable solution for adaptive robot control.

Several areas of the Central Nervous System (CNS) contribute to the temporal coordination implied in motor control such as the premotor cortex, the parietal cortex, the primary motor cortex, and the cerebellum [13], which stands out by its role in the integration, regulation, coordination of motor processes and more importantly, motor learning [14-17]. The cerebellum can be regarded as a separate area of the brain to which it is attached underneath the cerebral hemispheres, whose neural structure is highly regular in striking contrast to the cerebral cortex neural structure. This well-known structure makes it a suitable reference for the development of biologically plausible SNNs.

The depicted scenario yields several elements: 1) the cerebellum; a highly regular neural structure, thus, easy to computationally replicate to some extent, which is responsible for motor learning and coordination, 2) an artificial SNN incorporating a continuous learning process at its core that is able to mimic biological neurons and neural processing, and 3) hardware compliant robots lacking compliant control strategies. Here, we conjugate these three elements taking a holistic approach in tackling the HRI compliance problem.

Addressing this problem implies state-of-art challenges that we face along this work.

First, we need the cerebellar-like SNN to operate in RT. Spiking neural processing in RT is a highly demanding task in terms of computational cost. Considering that our computational resources are limited, there must be a trade-off between network size, neuron complexity, network topology and temporal output resolution, which determines, to certain extent, the motor control accuracy. We further developed our spiking neural simulator (EDLUT) to accommodate, for the first time, a RT cerebellar SNN consisting of ~62 K leaky integrated and fired (LIF) neurons with ~36.4 M synapses, 36 M of which are endowed with plasticity.

Second, we need to implement an effective RT dialogue between the network spike domain and sensorimotor analogue domain. In closed loop, the movements caused as a consequence of the sensory stimuli require that the SNN generating the motor commands receives an adequate driving input to generate an adequate motor output. This task is entrusted to the primary motor cortex (M1) which generates this input drive as a transformed version of the initial sensory signal [18]. Here, we emulate this M1 sensory transformation using a set of analogue-to-spike/spike-to-analogue modules compatible with Robot Operating System (ROS). These modules operate in RT without compromising motor accuracy.

Third, we need to cope with hardware/software compliance impositions. A compliant interaction with an unstructured environment [19] compels us to use a compliant robot (e.g. Baxter robot) in direct torque control. Using a compliant robot, such as Baxter, forces us to compensate, via the SNN controller, Baxter's loss in precision and lower capacity to exert a force due to its inner hardware compliance. We provided a compliant control in which a cerebellar-like SNN is able to continuously learn the minimal torque values needed to execute certain motor tasks in RT even under changing operational and ambient conditions, i.e. perturbation forces that continuously readjust their module and direction, human collisions, and interactions.

Finally, we need to assess the degree of goodness of the implemented solution. We have provided a compliant control in which a SNN is able to learn the adequate torque values in a safe manner. Furthermore, it is remarkable that our compliant control outperforms the accuracy achieved by the default factory-installed position control.

All in all, this work is the answer to overcome the technical difficulties aforementioned whose actual outcome provides us with a novel control strategy for hardware compliant robots based on a spiking cerebellar structure, which replicates the biological learning mechanisms involved in motor control.

## II. MATERIAL AND METHODS

### A. Benchmarking the Cerebellar Controller; Behavioural Tasks

We drew inspiration from the cerebellar role in motor control and movement coordination [15, 17] to implement a novel control strategy for hardware compliant robots. It is thus appropriate to evaluate our cerebellar-like model in the field of robot dynamics control in terms of performance under a set of different conditions. To this aim, we proposed a specific way of performing the experimental evaluation through two trajectory families.

1) On the one hand, we tested our cerebellar controller in reaching movements; that is, fast, ballistic arm

movements with bell-shaped velocity profiles, i.e. s-curve, towards a target point [20]. Arm reaching movements are primary used for characterising cerebellar pathologies in human motor control by measuring the time to target and precision to target. Arm dynamics control is critical due to the constraint at stake when moving masses. A single-joint limb movement in fast de/acceleration causes motion in all other limb joints thus arising interaction forces to be compensated by the cerebellum as well as our controller [21].

2) On the other hand, we tested our cerebellar controller facing a set of fast movements in smooth trajectories consisting of sinusoidal-like profiles for both position and velocity per joint. The end-effector shall describe either circular or eight-like Cartesian trajectories in the horizontal plane [22, 23]. These trajectories are well suited for revealing the complex dynamics of a 6 DoF robotic arm [24], including interaction forces to be compensated by the cerebellar controller [21].

These trajectory families were first designed in Cartesian space, providing 3D position and orientation for the end-effector, and then translated into joint space using MoveIt! software [25]. This offline process allowed the pre-computation of joint space trajectories that were later on used as cerebellar input. The circular trajectory in Cartesian space meets (1), whilst (2) describes the eight-like trajectory

$$\left.\begin{array}{l} x = R \cdot \cos(\Phi) \\ y = R \cdot \sin(\Phi) \\ z = \alpha \end{array}\right\} \Phi \in [0, 2 \cdot \pi]; \alpha = const \qquad (1)$$

$$\left.\begin{array}{l} x = 0.5 \cdot R \cdot \sin(2 \cdot \Phi) \\ y = R \cdot \cos(\Phi) \\ z = \alpha \end{array}\right\} \Phi \in [0, 2 \cdot \pi]; \alpha = const \qquad (2)$$

where $R$ denotes a 120 mm radius which is halved for the x coordinate in (2) to keep the eight-like trajectory within the working space limits of the robot. The $z = \alpha$ coordinate and the end-effector vertical orientation were kept constant to maintain the horizontal plane through the trajectories. Each trajectory lasted 2 seconds.

Once the translation from Cartesian (x, y, z) to joint space positions ($Q_{1 \text{ to } 6}$) was completed, the joint velocity profiles ($\dot{Q}_{1 \text{ to } 6}$) were obtained as the position derivative over time. Regarding the target-reaching task, the centre of the circle trajectory was the starting position. Eight different points along the circular trajectory perimeter constituted the reaching targets following an even distribution at every $\pi/4$ radians. As aforementioned, this task tested the controller through point-to-point multijoint movements with s-curve velocity profiles that provided fast acceleration/deceleration changes, i.e. ballistic movements. The subsequent high jerk values entailed high inertial forces to be compensated by the cerebellar controller. Each target-reaching movement lasted 2 seconds back and forth between the target and the central position, i.e. 1 second to reach the target and 1 second to go back to the central position. These three different behavioural tasks provided us with a varying context to test the cerebellar network. For every task, the cerebellar network acquired those motor commands needed to achieve the desired goal behaviour through learning. The learning process was accomplished through the repetition over time of a specified trajectory.

The performance evaluation was carried out comparing the goal and the actual behaviour, i.e. the desired and the actual joint positions. The average difference constitutes the position Mean Absolute Error (MAE), which is our performance evaluation metric following (3) and (4)

$$MAE_{joint} = \sum_{i=0}^{K} \left( Q_{i,desired} - Q_{i,actual} \right) \qquad (3)$$

$$MAE = \frac{\sum_{j=1}^{N} MAE_j}{N} \qquad (4)$$

where $K = 1000$ denotes the number of samples of the two second trajectories; and $N = 6$ is the number of joints. The MAE provided a numerical performance indicator for the quality of the cerebellar controller, thus allowing us to compare it against the default factory-installed position control.

### B. The Compliant Robot; the Baxter Robot

The Baxter robot®, manufactured by Rethink Robotics™ [26], is a collaborative robot consisting of two arms with seven DoF. Baxter implements torque control and it is inherently compliant thanks to its series elastic actuators (SEAs) [27]. These SEAs interpose a spring between the motor/gearing elements and the final motor output. These springs are deformable under human interaction and, therefore, a built-in mechanism that inherently allows for safe, compliant physical HRI

Prior to Baxter's hands-on testing, we used the simulated version of Baxter available in Gazebo as a safe environment to develop and test the robot-cerebellum interface [28]. This interface was developed using ROS to control both the simulated and real robot. ROS allowed sending motor commands (torque commands) to the robot and receiving sensorimotor information (joints positions and velocities) from the robot sensors [29]. The designed trajectories for our study involved the torque control of 6 DoF of one arm of the robot.

### C. Cerebellar Control Loop

The Baxter robot and the cerebellar network interconnection required the establishment of a dialogue in which the exchange of sensorimotor information modified the behaviour of one another. This dialogue was framed within a closed control loop with negative feedback. See Fig. 1 for a control loop overview.

The cerebellar-like spiking model (implemented in EDLUT, see below) acted as the controller and computed a motor command at each time step (2 ms) to achieve the goal behaviour. To this aim, the controller computed the neural activity using as input information the robot state, the ideal trajectory to be performed by the robot arm, and the instructive signal obtained. The robot state (actual position, $Q_a$, and actual velocity, $\dot{Q}_a$, per joint) was provided by Baxter's sensors and then mapped into control signals. The desired trajectory signals (position, $Q_d$, and velocity, $\dot{Q}_d$, per joint) were provided by a trajectory generator module representing the motor cortex and other motor areas. The instructive ε signals (one per joint) were obtained by comparison of the desired trajectory and the robot state signals. Once the cerebellar network computed a motor command, it was sent to the robot inducing movement to the arm. Consequently, the cerebellar network input sensory information was modified, thus, closing the loop. Cerebellar input and output signals were updated every 2 ms (500 Hz) guaranteeing low latency, a mandatory requirement for RT computation.

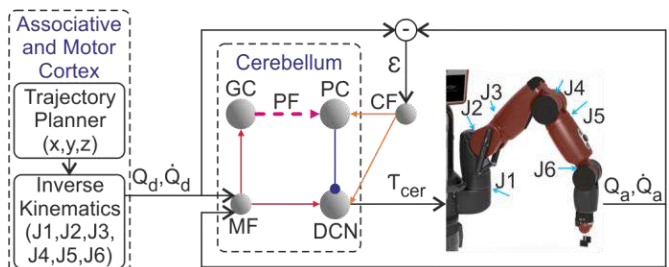

Fig. 1. Schematic of the Cerebellar closed-loop control. The Mossy fibres (MFs) convey the sensory signals, whilst the climbing fibres (CFs) convey the instructive signals, thus providing the inputs to the cerebellar network. The deep cerebellar nuclei (DCN) drive the cerebellar torque output commands. MFs project sensorimotor information onto granular cells (GCs) and DCN. GCs, in turn, project onto Purkinje cells (PCs) through parallel fibres (PFs). PCs also receive excitatory inputs from the CFs. Finally, DCN receive excitatory inputs from the MFs and CFs and inhibitory inputs from the PCs.

The cerebellar controller ran in EDLUT simulator [30-32]. EDLUT is mainly oriented to embodiment experimentation so that neural computation can be slowed down/speeded-up to cope with RT requirements imposed by a real body, e.g. humanoid robot [33, 34]. Regarding the theoretical concepts underpinning our cerebellar controller, please see our previous works [33, 35] on spike-analogue interfaces, [22, 36-38] on cerebellar learning, [37-40] on cerebellar granular layer, and [38, 41] on cerebellar control loops and neurorobotics.

### D. Cerebellar Controller; the Neural Network

The cerebellar network controller consisted of five neural layers: 1) Mossy fibres (MFs), 2) granule cells (GCs), 3) climbing fibres (CFs), 4) Purkinje cells (PCs), and 5) deep cerebellar nuclei (DCN) (see Fig. 2). The cerebellar network was in turn divided into six micro-complexes[42], each one focusing on controlling a different Baxter's joint.

The MFs constituted the input layer through which the input sensorimotor information was conveyed (actual and desired joint position and velocity trajectories translated into spiking patterns) towards the inner cerebellar network layers. These MFs projected excitatory afferents on both GCs and DCN. GCs, then, processed and re-coded this sensorimotor information in a sparse somatosensory neural activity that was later propagated by the parallel fibres (PFs) (i.e. excitatory GCs' axons) to the PCs. These PCs, in turn, correlated this somatosensory activity coming from PFs with the neural activity conveyed by the CFs (i.e. excitatory inferior olive, IO, axons). The CF neural activity, generated in the olivary system, represented the mismatch between the actual and desired trajectories per Baxter's joint and acted as an instructive signal. PCs underwent synaptic plasticity, that is, a supervised mechanism that correlated both PF and CF neural activities and adapted the PFs synaptic weight distribution accordingly. The cerebellar input-output response was adjusted and, therefore, the error movement minimised [43] in subsequent executions. Finally, the DCN closed the cerebellar loop via the excitatory synapses coming from MFs and CFs together with the inhibitory synapses from PCs.

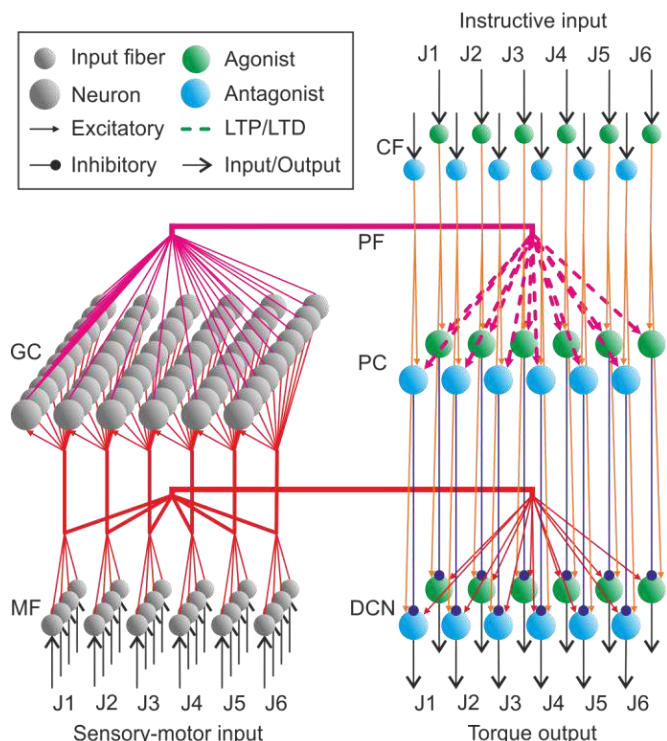

Fig. 2. Cerebellar scheme. Schematic representation of the main neural layers, cells, connections, and the plasticity site considered in the cerebellar model.

The DCN neural activity of each micro-complex ultimately drove each Baxter's joint by means of a spike-to-torque command transformation.

1) *MFs* (240) were modelled as input fibres able to propagate the sensorimotor information towards GCs and DCN at each simulation time step (2 ms). These 240 fibres were organised into six groups of 40 fibres each, i.e. one group per joint. Each MF group was in turn subdivided into four equal subgroups on which actual and desired joint positions and velocities were directly mapped. Only four non-overlapped MFs per group were active at each simulation time step representing the actual input neural state.

2) *GCs* (60,000) were modelled as LIF neurons emulating a state generator [40, 44, 45]. These 60,000 neurons were organised into six groups of 10,000 neurons each, i.e. one group per joint. Each GC received four input synapses [46] coming from each subgroup belonging to the very same MF group. The connectivity pattern between GC and MF groups was designed in a way that non-overlapped GC neural activation could univocally represent all possible MF neural input combinations. Importantly, this connectivity pattern facilitated the transformation of the sensorimotor neural information into a set of somatosensory neural activations that were easy to read out by the subsequent PC layer.

3) *CFs (600)* were modelled as input fibres able to propagate the instructive signal (mismatch between the actual and desired trajectories of each joint) towards PCs and DCN. These 600 fibres were organised into six micro-complexes of 100 neurons each, i.e. one per joint. Each micro-complex was also divided into two symmetrical subgroups, each one dedicated to control the clock/anticlockwise movement of the robot joint actuator (emulating the agonist/antagonist interplay in human muscles). A probabilistic Poisson process transformed the error obtained when comparing the actual and desired trajectories per joint into CF spiking neural activations. Each CF spike encoded well-timed information regarding the instantaneous error. The probabilistic spike sampling of the error ensured a proper representation of the whole error region over trials, whilst maintained the CF activity between 1 and 10 Hz per fibre (similar to electrophysiological data [47]). The error evolution could be sampled accurately even at such a low frequency [38, 48].

4) *PCs (600)* were modelled as LIF neurons. These 600 neurons were organised into six micro-complexes of 100 neurons each, i.e. one per joint. Each micro-complex was also divided into two symmetrical subgroups, each one dedicated to control the clock/anticlockwise movement of the robot joint actuator. Each PC was connected to all PFs, thus receiving the sensorimotor information concerning all joints at once. CFs and PCs were one-to-one connected maintaining the six-micro-complex architecture. Thus, each PC micro-complex received the same sensorimotor information via PFs, but a different instructive signal through its corresponding CFs micro-complex. Correlating these two different sources of neural information allows each PC micro-complex to adapt the cerebellar input-to-output response of each Baxter's joint via a plasticity mechanism that modified the overall PF synaptic weight distribution (see synaptic plasticity subsection).

5) *DCN (600)* were modelled as LIF neurons. These 600 neurons were organised into six micro-complexes of 100 cells neurons each, i.e. one per joint. Each micro-complex was also divided into two symmetrical subgroups, each one dedicated to control the clock/anticlockwise movement of the robot joint actuator. Each DCN cell was innervated by an inhibitory afferent from a PC and an excitatory afferent from the CF which simultaneously innervated the same PC. Each DCN cell also received excitatory projections from all MFs (which maintained the baseline DCN activity). This neural topology has been summarised in Table I.

TABLE I
NEURAL NETWORK TOPOLOGY

| Neurons | | Synapses | | | |
|---|---|---|---|---|---|
| Pre-synaptic cells | Post-synaptic cells | Number | Type | Initial weight (nS) | Weight range (nS) |
| 240 MFs | 60K GCs | 240K | AMPA | 0.18 | - |
| 240 MFs | 600 DCN | 144K | AMPA | 0.1 | - |
| 60K GCs | 600 PCs | 36M | AMPA | 1.6 | [0, 5] |
| 600 PCs | 600 DCN | 600 | GABA | 1.0 | - |
| 600 CFs | 600 PCs | 600 | AMPA | 0.0 | - |
| 600 CFs | 600 DCN | 600 | AMPA | 0.5 | - |
| 600 CFs | 600 DCN | 600 | NMDA | 0.25 | - |

The DCN neural activity was then transformed into an analogue torque command ($\tau_{cer}$) per micro-complex and then sent to Baxter's actuators. This spike to analogue conversion was computed at each time step, $T_{step} = 0.002$ s, using (5-7)

$$DCN_{j,i}(t) = \int_{t-T_{imp}}^{t} \delta_{DCN_{j,i}}(t) \cdot dt \qquad (5)$$

$$DCN_j(t) = \sum_{i=1}^{N=50} DCN_{j,i}(t) - \sum_{i=51}^{N=100} DCN_{j,i}(t) \qquad (6)$$

$$\tau_{cer,j}(t) = \frac{\alpha_j}{15} \cdot \sum_{x=1}^{15} DCN_{output}\left(t - (x-1) \cdot T_{step}\right) \qquad (7)$$

where $j \in \{1,6\}$ stands for the number of Baxter's joints; $i \in \{1,100\}$ defines the DCN tag number within the micro-complex related to joint $j$ (first 50 DCN cells encoding the agonist movement whereas last 50 DCN cells encoding the antagonist movement); and $\delta(t)$ stands for the Dirac delta function representing a spike event.

The spike to analogue conversion in (5) and (6) was then convolved with a mean filter (7) acting as a DCN activity eligibility trace; that is, a temporary record of the occurrence of DCN previous spike events. The fifteen-taps mean filter helped us to emulate the low-pass filter behaviour of muscles. The final torque output per joint was finally modulated by a factor $\alpha$ to adequate the normalised DCN output to the joint relative position, orientation and mass; $\alpha_j = (0.75, 1.0, 0.375, 0.5, 0.05, 0.05)$ N·m/spike.

### E. Spiking Neuron Models

The cerebellar neural network consisted of LIF neurons [49] due to their minimal computational cost in spike generation and processing, a key factor in RT computation. Our LIF neurons only elicited a spike once their corresponding membrane potential reached a certain threshold and, immediately after, their membrane potentials were reset. The LIF neural dynamics was just defined by its membrane potential and its excitatory (AMPA and NMDA) and inhibitory (GABA) chemical conductances as follows

$$C_m \cdot \frac{dV}{dt} = I_{internal} + I_{external} \qquad (8)$$

$$I_{internal} = -g_l \cdot (V + E_L) \qquad (9)$$

$$I_{external} = -\left(g_{AMPA}(t) + g_{NMDA}(t) \cdot g_{NMDA\_INF}\right) \cdot (V - E_{AMPA}) - g_{GABA}(t) \cdot (V - E_{GABA}) \qquad (10)$$

$$g_{AMPA}(t) = g_{AMPA}(t_0) \cdot e^{\frac{t-t_0}{\tau_{AMPA}}} + \sum_{i=1}^{N} \delta_{AMPA_i}(t) \cdot w_i \qquad (11)$$

$$g_{NMDA}(t) = g_{NMDA}(t_0) \cdot e^{\frac{t-t_0}{\tau_{NMDA}}} + \sum_{i=1}^{N} \delta_{NMDA_i}(t) \cdot w_i \qquad (12)$$

$$g_{GABA}(t) = g_{GABA}(t_0) \cdot e^{\frac{t-t_0}{\tau_{GABA}}} + \sum_{i=1}^{N} \delta_{GABA_i}(t) \cdot w_i \qquad (13)$$

$$g_{NMDA\_INF} = \frac{1}{1 + exp(62 \cdot V) \cdot \frac{1.2}{3.57}} \qquad (14)$$

where $C_m$ denotes the membrane capacitance; $V$ is the membrane potential; $I_{internal}$ is the internal current and $I_{external}$ is the external current. $E_L$ is the resting potential and $g_L$ the conductance responsible for the passive decay term towards the resting potential. Conductances $g_{AMPA}$, $g_{NMDA}$ and $g_{GABA}$ integrate all the contributions received by each receptor type (AMPA, NMDA, GABA) through individual synapses, being $g_{NMDA\_INF}$ the NMDA activation channel. These conductances were defined as decaying exponential functions [30, 49] where their values were directly incremented proportionally to the synaptic weights ($w_i$) upon each presynaptic spike arrival (Dirac delta functions). When the membrane potential reached a threshold ($V_{thr}$), it was then reset to $E_L$ during the refractory period ($T_{ref}$). The configuration parameters for the three neurons modelled are shown in Table II.

### F. Synaptic Plasticity

The adaptive motor process of the cerebellar network was implemented through a STDP mechanism located at PF-PC synapses. This STDP mechanism balanced long-term potentiation (LTP) and long-term depression (LTD) at PC synaptic level as follows

$$LTP \Delta w_{PF_{j}-PC_i}(t) = \alpha \cdot \delta_{PFspike}(t) \cdot dt \qquad (15)$$

$$LTD \Delta w_{PF_{j}-PC_i}(t) = \beta \cdot \int_{-\infty}^{t_{CFspike}} k(t - t_{CFspike}) \cdot \delta_{PFspike}(t) \cdot dt \qquad (16)$$

where $\Delta W_{PF_{j}-PC_i}(t)$ denotes the synaptic weight change between the $j^{th}$ PF and the target $i^{th}$ PC; $\alpha = 0.002$ nS is the synaptic efficacy increment; $\delta_{PF}$ is the Dirac delta function corresponding to an afferent spike from a PF; $\beta = -0.001$ nS is the synaptic efficacy decrement; and the kernel function $k(x)$ is defined as

$$k(x) = \begin{cases} \dfrac{-(x+d_k)}{\tau_{LTD} - d_k} \cdot e^{\frac{x+d_k}{\tau_{LTD}-d_k}+1} & if\ x < -d_k \\ 0 & if\ x \geq -d_k \end{cases} \qquad (17)$$

where $\tau_{LTD} = 100$ ms is the time constant that is aligned with the biological sensorimotor pathway delay (~100 ms), the time period elapsed from the sensory information reception to, information transmission along nerve fibres, neural processing time responses and the final motor output response [50]. $d_k = 0.07$ s allows for the adjustment of the kernel width. The kernel maximum value ($k(x) = 1$) is obtained when $x = -\tau_{LTD}$, and zero or close to zero when $x > -d_k$ or $x < -\tau_{LTD} - 10 \cdot (\tau_{LTD} - d_k)$. The STDP rule defined by (15-17) caused a fixed synaptic efficacy increment (LTP) each time a spike arrived through the PFs to the target PC and a variable synaptic efficacy decrement (LTD) each time a spike arrived through a CF to the target PC. The amount of synaptic decrement depended on the activity arrived through the PFs prior to the CF spike. Both activities were convolved using the integrative kernel defined in (17) and were multiplied by the synaptic decrement $\beta$. The effect on the presynaptic spikes arriving through PFs was maximal during the 100 ms time window ($\tau_{LTD} = 100$ ms) before the postsynaptic CF spike arrival, thus accounting for the sensorimotor pathway delay [38, 41, 51].

This STDP mechanism correlated the neural activity patterns coming through the PFs towards PCs with the instructive signals coming from CFs towards PCs. This correlation process at PC level identified certain PF activity patterns codifying certain sensorimotor information and, consequently, diminished the PC output activity by a PF-PC synaptic weight reduction. A reduction on the PC activation caused a subsequent reduction on the PC inhibitory action over the target DCN. Conversely, in the absence of any correlation, the STDP mechanism increased the PC output activity by a PF-PC synaptic weight potentiation. Since the DCN were driven by a near constant baseline MF activation, a lack of PC inhibitory action would cause an increasing DCN activity whereas an incremental PC inhibitory action would do otherwise. Well-timed sequences of increasing/decreasing levels of DCN activation during the learning acquisition process ultimately shaped the cerebellar output activity and diminished the overall error.

### G. ROS modules implementation

The control loop consisted of three main elements: 1) trajectory generator, 2) cerebellar controller, and 3) Baxter robot. The implementation and communication amongst these three elements were developed using ROS, allowing modularity. Fig. 3 depicts the control loop diagram in which each block defines a ROS module and each black arrow represents a ROS topic that establishes the communication between ROS modules exchanging either analogue signals or spike trains.

This control loop was designed accounting for the sensorimotor pathway delay (~100 ms) [52]. The 100 ms delay comprised the efferent ($\delta_x = 50$ ms) and afferent ($\delta_a = 50$ ms) pathway delays (Fig. 3 dashed red arrows). A motor command originated at time $t$ on the cerebellum was applied by the robot actuators at time $t + \delta_e$ and its effect sensed back at the cerebellar network at time $t + \delta_e + \delta_a$. The cerebellar plasticity mechanism described in (15-17) compensated for this sensorimotor delay.

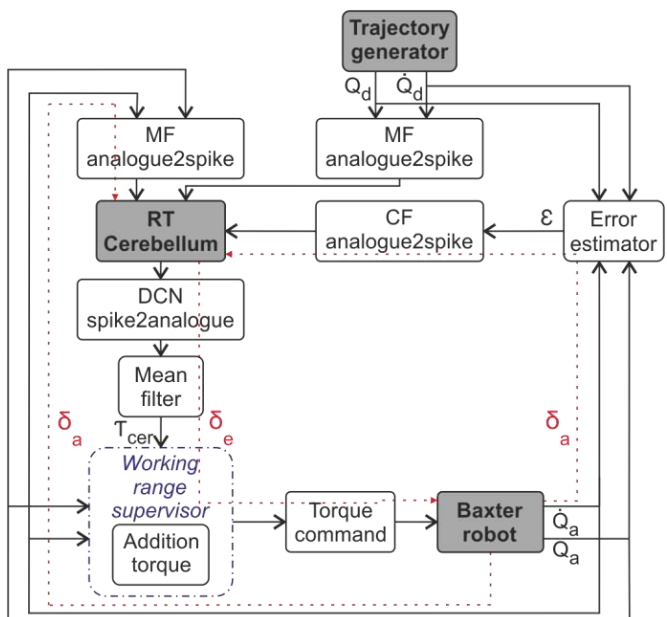

Fig. 3.  Detailed cerebellar closed-loop control scheme.

The control loop assisted the cerebellar controller in the generation of the torque commands able to minimise the mismatch between the reference signal (desired joint position and velocity) and the robot state (actual joint position and velocity). To that, different ROS modules were implemented:

1) The trajectory generator module generated the desired trajectory signals, whilst the Baxter robot generated the actual state signals and executed the motor commands.

2) The RT cerebellum module accommodated the cerebellar controller implemented in EDLUT (imported in ROS as a C++ library). This module received, computed, extracted and propagated the neural activity towards the next ROS module.

3) Desired, actual and instructive signals needed to be transformed into spike trains that the cerebellum could process. The MF and CF analogue2spike modules carried out this transformation.

4) The error estimator module provided the cerebellar controller with the instructive signal needed for neural adaptation. The error estimator module required comparing desired and actual trajectories.

5) The cerebellar output spiking signals needed to be transformed into analogue commands that Baxter robot could process. The DCN spike2analogue module, using (5) and (6) transformed the spike trains into torque commands, which were lately smoothed out by the mean filter module using (7).

6) The torque command module closed the loop sending the torque commands obtained from the mean filter module to the Baxter robot.

7) The supervisor module was implemented as a safe mechanism mimicking mechanical brakes. A supervisor module maintained Baxter within a safe working range during the first stages of neural adaptation. Only at the event of any of the joints getting outside its working range, the supervisor module added a corrective torque value to the cerebellar torque command to prevent damages.

All modules were synchronised thanks to a reference time signal extracted from Baxter's internal clock running under the Network Time Protocol (NTP), ensuring RT [53]. Each event, i.e. analogue signal or spike train, generated on a ROS module carried a time stamp indicating the event processing time to the subsequent module. Each target module incorporated an input buffer in which events were stored for later synchronous processing according to their time stamps. The RT cerebellum module, however, allowed for asynchronous processing of the events stored at its input/output activity buffers thanks to the RT mechanism incorporated in EDLUT [34]. On the event of empty input buffers, the neural simulation was halted. On the event of an almost empty output buffer, the neural simulation was speeded-up (see [34] for an in-depth review on RT neural simulation). Hence, the RT cerebellum module could deal with neural activity volleys encountered during the cerebellar simulation that could not be processed synchronously.

III. RESULTS

We tested our cerebellar-like controller under different conditions, i.e. behavioural tasks, considering the default factory-installed position control mechanism as a performance baseline to validate the results. The aforementioned circular, eight-like and target reaching trajectories constituted our cerebellar benchmarking, which was completed with a set of interactions in an unstructured environment to test compliance.

A. Circular Trajectory

This first behavioural task consisted in following a 120 mm radius circular path in the horizontal plane ($xy$) repeated over time to facilitate learning and adaptation, each trial having a time duration of 2 seconds. The STDP mechanism governing the learning process modulated the cerebellar output (see Methods), driving the robot's behaviour towards the goal. The behavioural evolution through time is illustrated in Fig. 4. Three snapshots were taken at three different moments of the cerebellar learning process: initial, intermediate, and final stage.

1) Initial learning stage: The cerebellar-model started learning from scratch. At an initial learning stage [Fig. 4 left column)] the synaptic adaptation mechanism at PF-PC synapses that correlated the somatosensory information with the CF instructive signal was not effectively deployed yet. Thus, the inhibitory action from PCs onto DCN was of marginal utility; making the DCN output activity saturated as it solely responded to the excitation coming from MF and CF afferents [Fig. 4 (a), first row]. Consequently, the corresponding initial torque commands [Fig. 4 (a), second row] were far from leading the robot towards the desired goal [Fig. 4 (a), third row;

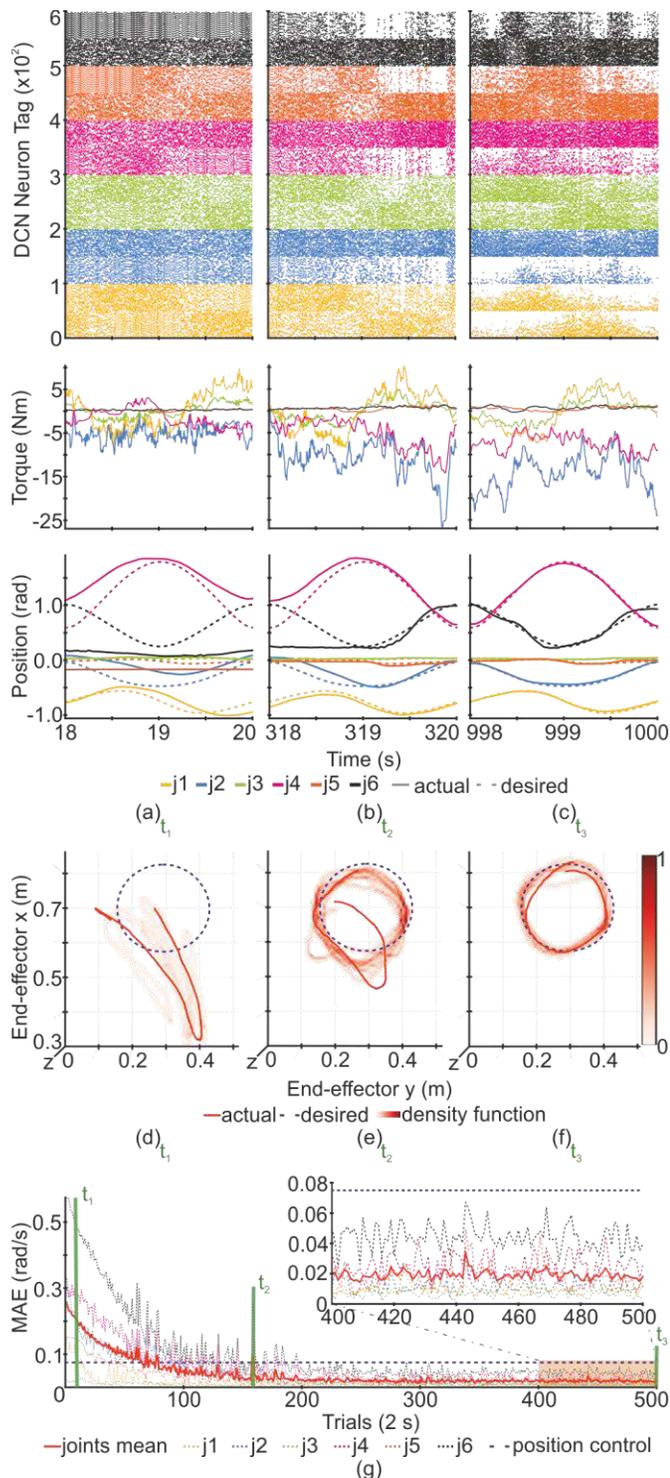

Fig. 4. Behavioural evolution through circular trajectory trials (2 s). (a) Initial learning stage ($t_1$=18-20 s). (b) Intermediate learning stage ($t_2$=318-320 s). (c) Final learning stage ($t_3$=998-1000 s). The first row depicts the cerebellar output activity (DCN layer), whereas the second row shows its analogue conversion into torque commands. The third row illustrates the desired vs. actual trajectory per joint. (d), (e), and (f) reveal the desired vs. actual trajectory of the end-effector in Cartesian space at $t_1$, $t_2$, and $t_3$ respectively, along with the density functions corresponding to the performed trajectories of the prior 10 trials. (g) Represents the position Mean Absolute Error (MAE) per trial through the learning process. Comparison of the MAE of each joint and the mean of all joints with the default factory-installed position control baseline performance.

and (d)]. As depicted in Fig. 4 (d), the density function generated from 10 trials before t1 snapshot [Fig. 4 (left column)] reveals that the robot was still exploring the working area, performing low consistent, dispersed movements.

2) Intermediate learning stage: At an intermediate learning stage [Fig. 4 (central column)], the synaptic adaptation allowed the recognition of some somatosensory patterns at the PCs, which was reflected in an emerging differentiated DCN activity between agonist and antagonist subgroups at each micro-complex [Fig. 4 (b), first row]. Consequently, the robot's behaviour began getting closer to the desired goal [Fig. 4 (b), third row; and (e)].

3) Final learning stage: Once the learning process reached advanced stages [Fig. 4 (right column)] the robot executed the desired trajectory with minimal error. The agonist/antagonist DCN activity was clearly differentiated at each micro-complex [Fig. 4 (c), first row], and translated into the required torque commands via a spike-to-analogue conversion (see Methods). The synaptic adaptation process was reflected in a clear evolution of the torque values compared to previous stages, directly affecting the robot output behaviour. All joints closely followed the desired trajectory at this stage [Fig. 4 (c), third row] and, consequently, the end-effector barely missed at describing the desired circular path [Fig. 4 (f)], having a consistent behaviour around the goal trajectory over trials.

The overall performance through the learning process is depicted in Fig. 4 (g); illustrating how the cerebellar-like controller performance was improved as adaptation and learning were fulfilled. MAE evolution indicates that the cerebellar controller needed about 300 trials (i.e. 600 seconds) to converge, outperforming the accuracy of the default factory-installed position control baseline ($0.019 \pm 0.003$ vs. $0.077 \pm 0.0004$, Table III).

### B. Eight-like Trajectory

The eight-like trajectory was concentric to the previously discussed circle-shaped; it had a "radius" of 120 mm and each trial lasted 2 seconds. In terms of robot dynamics, the eight-like trajectory was more demanding than the circular trajectory, as faster and steeper changes in velocity module and direction were required for trajectory completion[24]. Nonetheless, the obtained results were equally satisfying (see Table III).

1) Initial learning stage: At an early learning stage [Fig. 5 (left column)] the robot's behaviour was clearly far from the desired goal. DCN activity at this stage responded exclusively to the excitatory drive from MF-DCN and CF-DCN afferents, thus, it was saturated [Fig. 5 (a), first row]. The MAE value was high (0.165) and the performed trajectory was far from the goal [Fig. 5 (a), third row; (d), and (g)].

2) As learning progressed, the PF-PC synaptic adaptation mechanism begun shaping the DCN activity causing an incipient neural activity differentiation between agonist and antagonist micro-complexes [Fig. 5 (b), first row]. In

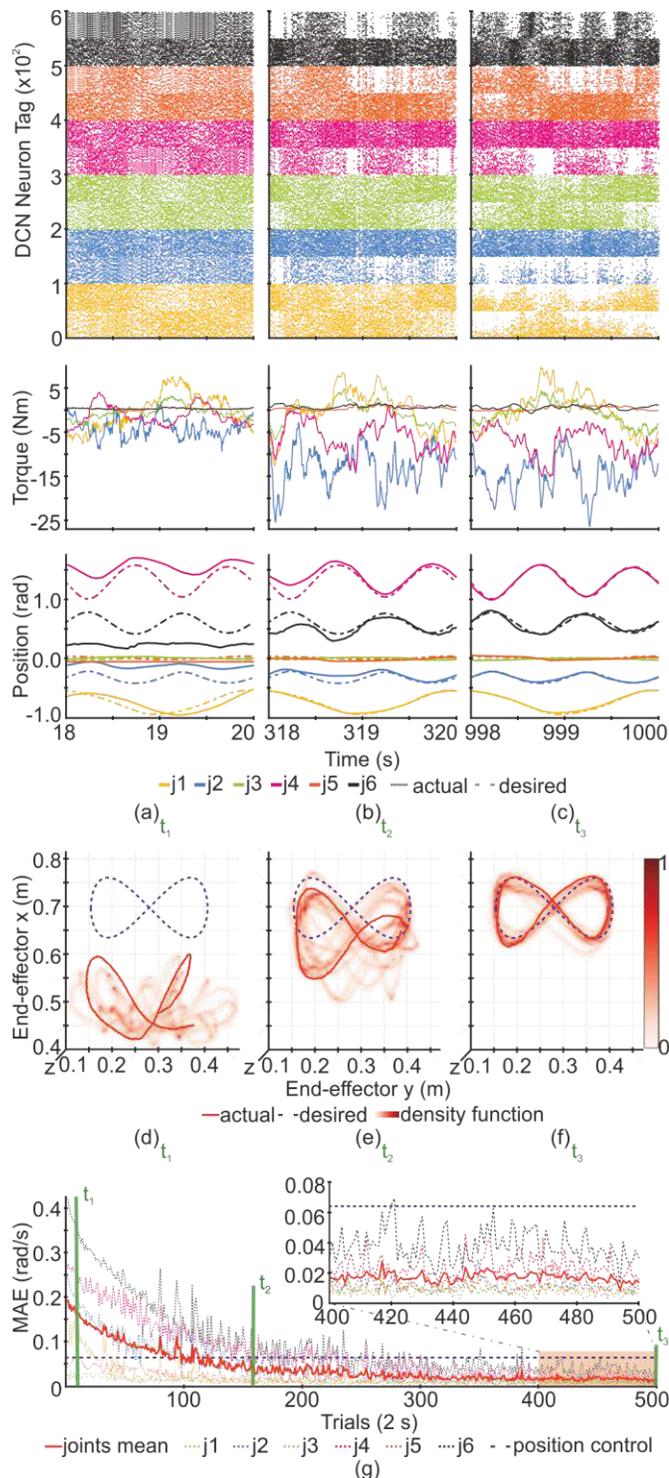

Fig. 5. Behavioural evolution through eight-like trajectory trials (2 s). (a) Initial learning stage ($t_1$=18-20 s). (b) Intermediate learning stage ($t_2$=318-320 s). (c) Final learning stage ($t_3$=998-1000 s). The first row depicts the cerebellar output activity (DCN layer), whereas the second row shows its analogue conversion into torque commands. The third row illustrates the desired vs. actual trajectory per joint. (d), (e), and (f) reveal the desired vs. actual trajectory of the end-effector in Cartesian space at $t_1$, $t_2$, and $t_3$ respectively. Also the density functions corresponding to the prior 10 trials are depicted. (g) Represents the position Mean Absolute Error (MAE) per trial through the learning process. The MAE of each joint is illustrated as well as the average MAE of all joints, completed with the default factory-installed position control baseline performance.

consequence, the corresponding torque values significantly differed from those of early stages [Fig. 5

### TABLE III
CIRCULAR AND EIGHT-LIKE TRAJECTORIES: LEARNING STAGES MAE

| | | Cerebellar torque control (trials) | | | Position control (trials) |
|---|---|---|---|---|---|
| | | [0-100] | [100-200] | [400-500] | [0-500] |
| MAE | ○ | 0.115 ± 0.055 | 0.036 ± 0.013 | 0.019 ± 0.003 | 0.077 ± 0.0004 |
| | ∞ | 0.111 ± 0.034 | 0.046 ± 0.013 | 0.017 ± 0.003 | 0.063 ± 0.0003 |

(b), second row], and the robot's behaviour began getting closer to the desired one [Fig. 5 (b), third row; and (e)].

3) Finally, once learning was fully deployed the robot behaved as desired [Fig. 5 (c), third row; and (f)]. The DCN activity was clearly sculpted to produce the needed torque commands to perform the desired trajectory [Fig. 5 (c)], maintaining a stable behaviour over trials (0.017 ± 0.003).

The greater difficulty of the eight-like trajectory was noted in a lower convergence speed for the cerebellar-like controller to reach a stable behaviour (Table III shows a slower MAE convergence speed than the circular trajectory). However, the final performance accuracy obtained also outperformed the default factory-installed position control baseline (0.017 ± 0.003 vs. 0.063 ± 0.0003).

### C. Target Reaching

This task consisted of eight different reaching movements, sharing the same starting point. The challenge lied in the high speed of the movements and the randomness in the order of trials (transitions between the eight reaching movements were stochastic). The growth in complexity for the cerebellar controller was illustrated by a lower MAE convergence speed entailing higher standard deviation values inter trials and the need of more trials to reach stability than the two previous behavioural tasks (Table IV). Nevertheless, the cerebellar-like controller was able to perform these ballistic movements, improving its performance through learning and reaching again better accuracy than the default factory-installed position control mechanism [Fig. 6] (0.019 ± 0.006 vs. 0.026 ± 0.006).

Therefore, not only the cerebellar-like controller was able to perform accurate smooth trajectories but also fast-ballistic movements.

### TABLE IV
TARGET REACHING: LEARNING STAGES MAE

| | Cerebellar torque control (trials) | | | Position control (trials) |
|---|---|---|---|---|
| | [0-100] | [300-400] | [900-1000] | [0-1000] |
| MAE | 0.155 ± 0.050 | 0.043 ± 0.024 | 0.019 ± 0.006 | 0.026 ± 0.006 |

### D. Unstructured interactions

Aiming at testing the compliance of the cerebellar controller, we tested its response in an unstructured environment. Whilst performing the circular trajectory, some interactions were undertaken [Fig. 7]. First, the dynamics of the robotic arm was modified in two different ways: i) By adding a 0,5kg payload to the end-effector attached to a rod, mimicking a pseudo "conical pendulum". The tension force of the rod acting on the robot varied with the alignment between the payload and the end-effector. ii) By attaching an elastic band to apply an elastic force that tried to return the band to its natural length. In both cases, the cerebellar-like controller successfully adapted to the new context after a learning period.

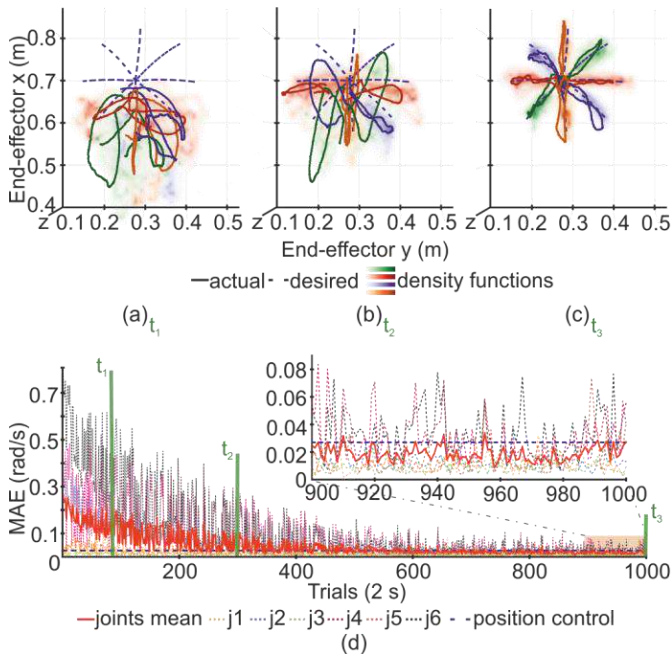

(a) $t_1$          (b) $t_2$          (c) $t_3$

(d)

Fig. 6.  Behavioural evolution through target reaching trials (2 s). Each trial consisted of one of the eight possible tasks. (a) Initial learning stage ($t_1$=158-160 s). (b) Intermediate learning stage ($t_2$=598-600 s). (c) Final learning stage ($t_3$=1998-2000 s). (a), (b), and (c) depict the last performed trajectory for each of the eight possibilities in Cartesian space prior to $t_1$, $t_2$, and $t_3$ respectively. The density functions reveal the end-effector behaviour over the last 80 trials, grouping the eight possible tasks by trajectory direction. (d) Represents the position Mean Absolute Error (MAE) per trial through the learning process. The MAE of each joint is illustrated as well as the mean MAE of all joints. High standard deviation values reflect how some reaching movements were more demanding than others. The position control baseline is the average MAE of the default factory-installed under the same stochastic distribution over trials.

Subsequently, human interactions were performed: i) A human was able to move the robotic arm by applying an extremely low force (i.e. one-finger push). ii) A human grabbed the robotic arm and moved it through the working space with no opposition from the robot. iii) A human got in the way of the robotic arm trajectory with no risk for injury.

These results allow us to confirm that the cerebellar-like controller was able to accurately perform the desired trajectories, no matter the dynamics modifications; and guaranteed a safe human-robot interaction as no damages were suffered when interrupting the robot's task, at either human or robot side.

Four movies are included as supplementary material to fully illustrate the cerebellar learning and adaptation process. The target reaching, eight-like, and circular trajectory movies show from up to down and left to right the following clips, all of them playing synchronised RT information: i) a frontal shot of the robot performing the trajectory; ii) the evolution of the position MAE per trial; iii) a nadir shot of the robot performing the trajectory; iv) the trajectory being performed by the end-effector in Cartesian space; v) the cerebellar output activity (DCN layer spikes); vi) the corresponding torque commands obtained from the spike-to-analogue conversion of the DCN activity. Different cuts corresponding to an initial, intermediate, and final learning stage verify the behavioural evolution.

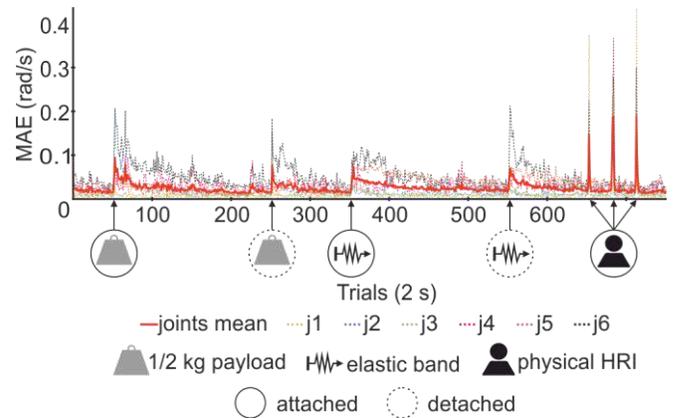

Fig. 7.  Performance in an unstructured environment. Whilst performing the already learnt circular trajectory a set of unstructured interactions were undertaken: i) A ½ kg payload was attached to the end-effector and later on detached. ii) An elastic band was attached to the end-effector and later on detached. iii) A series of physical Human-Robot interactions. The figure depicts the position MAE through trials as interactions are undertaken, illustrating the cerebellar adaptation to unknown scenarios.

Finally, the unstructured environment movie shows the cerebellar adaptation and, therefore, robot adaptation, to unknown, unstructured scenarios; thus, proving compliance.

## IV. CONCLUSION

Physical HRI implies controlling nonlinearities at the robotic end, thus demanding adaptive control. In this work, taking biology as an inspiration, we expand the family of RT adaptive robot controllers beyond machine learning [54], fuzzy logic [55, 56] and ANNs [9, 57] solutions. We present a novel biologically plausible motor control architecture with a cerebellar-like SNN controller at its core that is able to drive a 6 DoF robot via torque commands in RT.

The intrinsic characteristics of SNNs, i.e. timing codification of evolving sensorimotor states, make them an appealing approach for motor control architectures [11, 12]. However, computational cost has been the major drawback for implementing RT SNN controllers [58]; constraining their applicability to little versatile hardware solutions [59, 60], simulated scenarios [56, 58], or RT with low resolution control signals [61].

Here, this main issue has been overcome; a ~62 k neuron sized SNN, endowed with plasticity (36M plastic synapses), has been proven a valid RT robot controller. The implemented cerebellar plasticity mechanism (STDP) turns dispensable the availability of a detailed dynamic model of the robot. The cerebellar-like SNN is able to self-adapt and learn from scratch to control a given robot, making unnecessary any prior dynamics knowledge. Thus, the complexity of modelling nonlinear systems is tackled, and this SNN controller constitutes a plausible solution to control not only our Baxter robot, but any torque controlled robot. Previously achieved SNN position control [61, 62] does not provide compliance as physical perturbations/interactions are not supported; hence, the importance of reliable torque control towards achieving safe physical HRI.

The variety of demanding dynamic tasks in terms of control requirements here accomplished proves our SNN cerebellar-like controller a valid solution. Our SNN controller succeeded in terms of position accuracy, high-speed movements and compliance since the baseline performance (i.e. default factory-installed position control) was utterly improved in all the experimental behavioural tasks.

The development of biologically plausible controllers appears as a driving force for the evolution of robotics towards more advanced, intelligent, bio-inspired and compliant robots. Furthermore, the embodiment of biologically accurate artificial neural networks implies a great opportunity for neuroscience studies. These neural network models can be computationally simulated under different biologically relevant tasks to give a consistent idea about how the CNS neural network may operate.

## Acknowledgment

The authors would like to thank Dr. F. Barranco and Dr. R. R. Carrillo, for their research inputs.

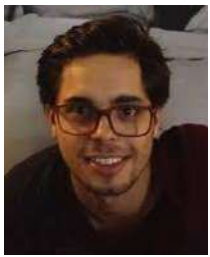

**Ignacio Abadía** received his B.Sc. and M.Sc. degrees in telecom engineering and secondary education teaching from the U. Granada (Spain) in 2015 and 2016 respectively. In 2018, he joined the Applied Computational Neuroscience Research Group of the U. Granada (ACN-UGR) awarded with a Young Researchers fellowship. His main research interests include neuromorphic engineering, spiking neural networks, brain-computer interfaces and motor control.

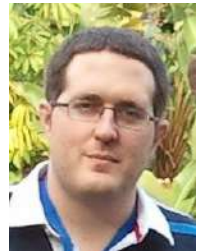

**Francisco Naveros** received two M.Sc. degrees, in telecommunication, and in computer science and networks in 2011, 2012 respectively. He also holds a Ph.D. degree in computational neuroscience from the U. Granada (Spain), 2017. He has been a postdoctoral researcher since 2017 at ACN-UGR. He is the author of 8 articles. His main research interests include biologically processing control schemes, parallel and real-time spiking neural network and lightweight robots.

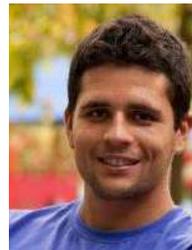

**Jesús A. Garrido** earned his M.Sc. degree in computer sciences in 2006 and his M.Sc. and PhD degree in computer engineering and networks in 2007 and 2011 respectively, all from the U. Granada (Spain). From 2012 to 2015, he joined the Brain and Behavioral Science department at U. Pavia (Italy) under supervision of Prof. D'Angelo. In 2015, he was awarded with a Young Researchers Fellowship by U. Granada. From 2016 to 2019, he obtained an IF Marie Curie Post-Doc Fellowship from the EU in the ACN-UGR. He is the author of more than 25 articles. His main research interests include cerebellar information processing and learning, motor control, neuromorphic engineering, and spiking neural networks

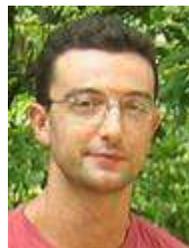

**Eduardo Ros** received his M.Sc. and Ph.D. degrees in physics and computational neuroscience from the U. Granada (Spain) in 1992 and 1997 respectively. He is currently Full Professor in the Dept. of Computer Architecture and Technology of U. Granada. He is the head of the ACN-UGR group. He is the author of more than 85 scientific articles. His main research interests include bio-inspired processing, neuromorphic engineering, spiking neural networks and computational neuroscience.

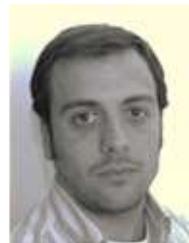

**Niceto R. Luque** was awarded his M.Sc. and Ph.D. degrees in computer science and networks from U. Granada (Spain) in 2007 and 2013 respectively. He also received a B.Sc in electronics and a M.Sc. in automatics and industrial electronics from U. Córdoba (Spain) in 2003 and 2006, respectively. From 2015 to 2017, he obtained an IF Marie Curie fellowship from the EU in Dr. Arleo's lab in Paris. In 2018 he obtained a Juan de la Cierva Incorporation Post-Doc fellowship from the Spanish Government in the ACN-UGR. He is the author of more than 20 articles. His main research interests include biologically processing control, spiking neural networks and ageing.

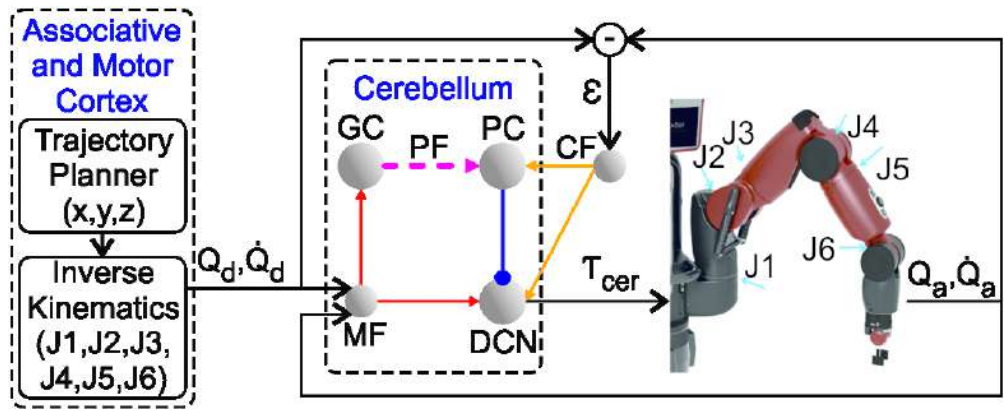

Fig. 1.  Schematic of the Cerebellar closed-loop control. The Mossy fibres (MFs) convey the sensory signals, whilst the climbing fibres (CFs) convey the instructive signals, thus providing the inputs to the cerebellar network. The deep cerebellar nuclei (DCN) drive the cerebellar torque output commands. MFs project sensorimotor information onto granular cells (GCs) and DCN. GCs, in turn, project onto Purkinje cells (PCs) through parallel fibres (PFs). PCs also receive excitatory inputs from the CFs. Finally, DCN receive excitatory inputs from the MFs and CFs and inhibitory inputs from the PCs.

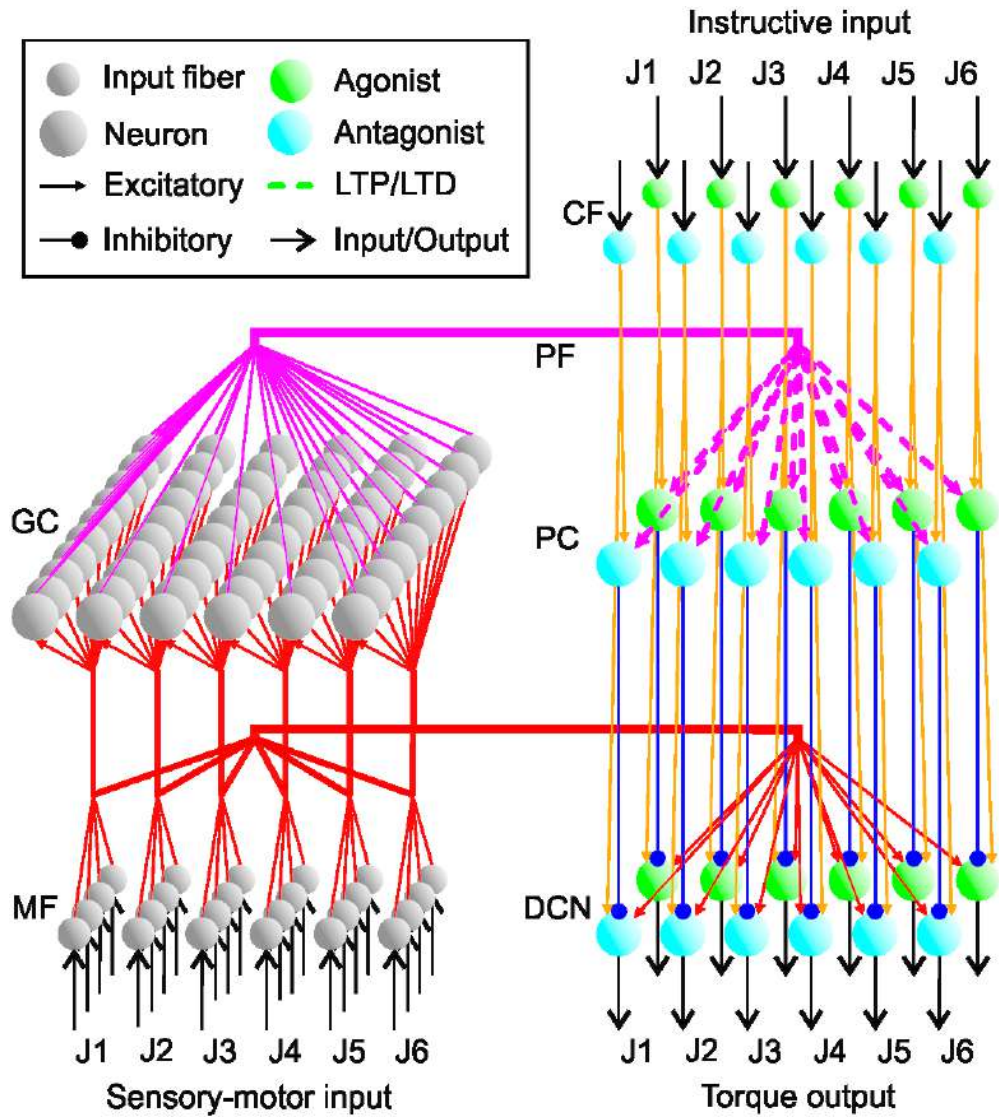

Fig. 2. Cerebellar scheme. Schematic representation of the main neural layers, cells, connections, and the plasticity site considered in the cerebellar model.

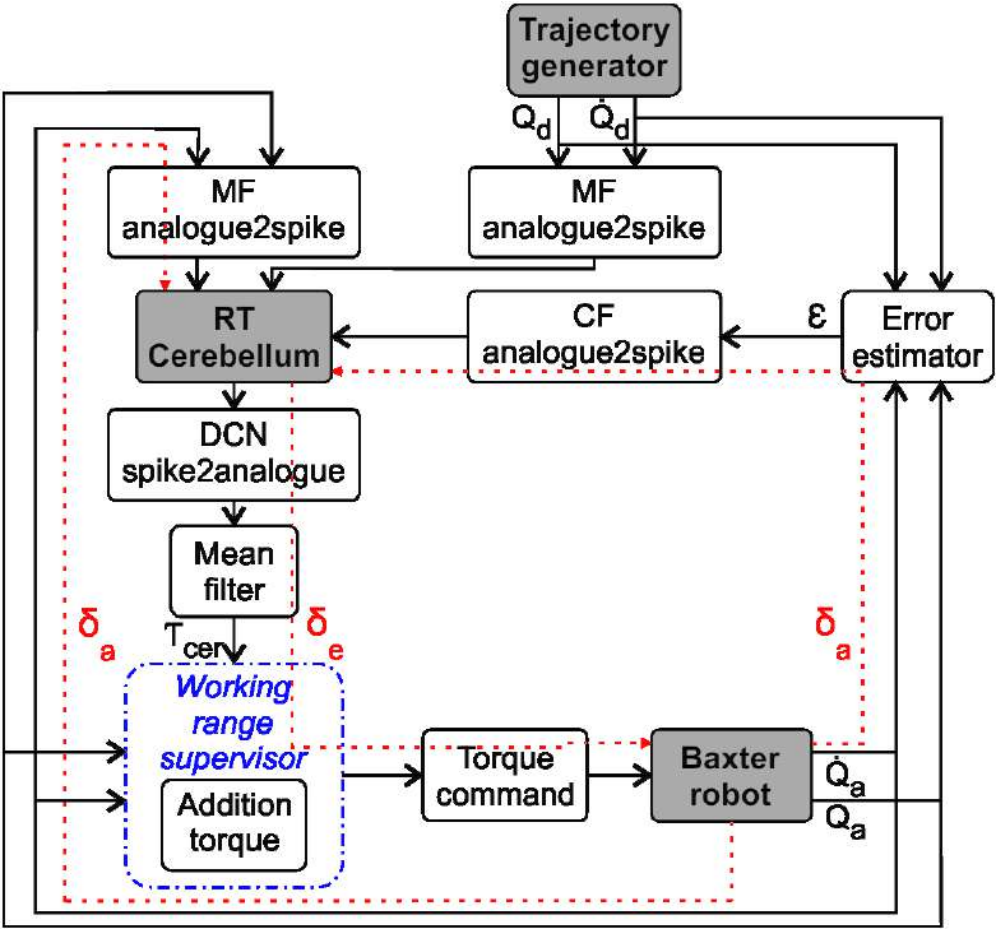

Fig. 3.  Detailed cerebellar closed-loop control scheme.

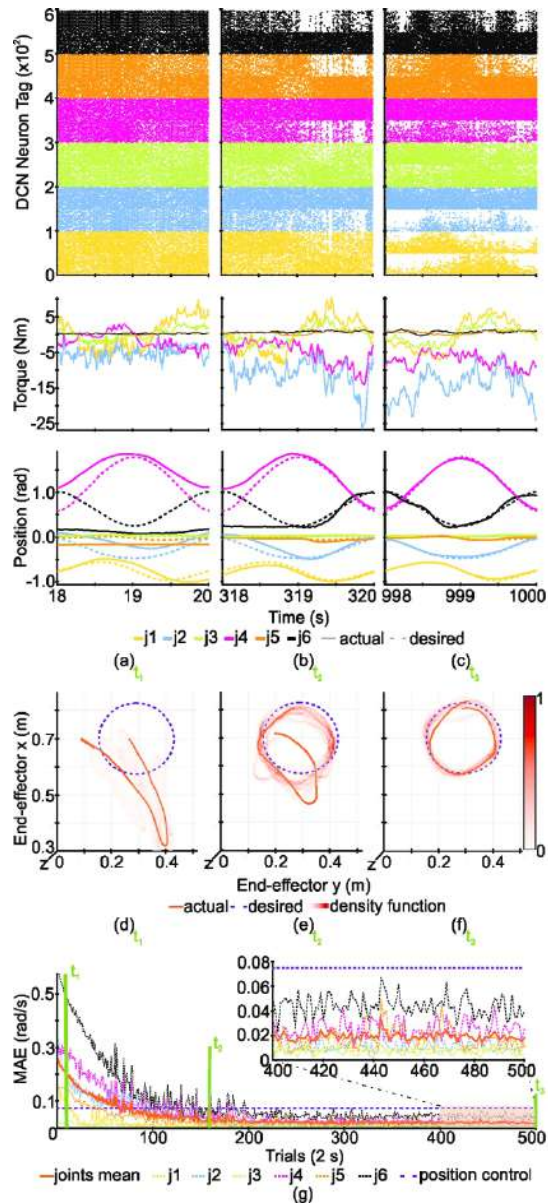

Fig. 4. Behavioural evolution through circular trajectory trials (2 s). (a) Initial learning stage (t1=18-20 s). (b) Intermediate learning stage (t2=318-320 s). (c) Final learning stage (t3=998-1000 s). The first row depicts the cerebellar output activity (DCN layer), whereas the second row shows its analogue conversion into torque commands. The third row illustrates the desired vs. actual trajectory per joint. (d), (e), and (f) reveal the desired vs. actual trajectory of the end-effector in Cartesian space at t1, t2, and t3 respectively, along with the density functions corresponding to the performed trajectories of the prior 10 trials. (g) Represents the position Mean Absolute Error (MAE) per trial through the learning process. Comparison of the MAE of each joint and the mean of all joints with the default factory-installed position control baseline performance.

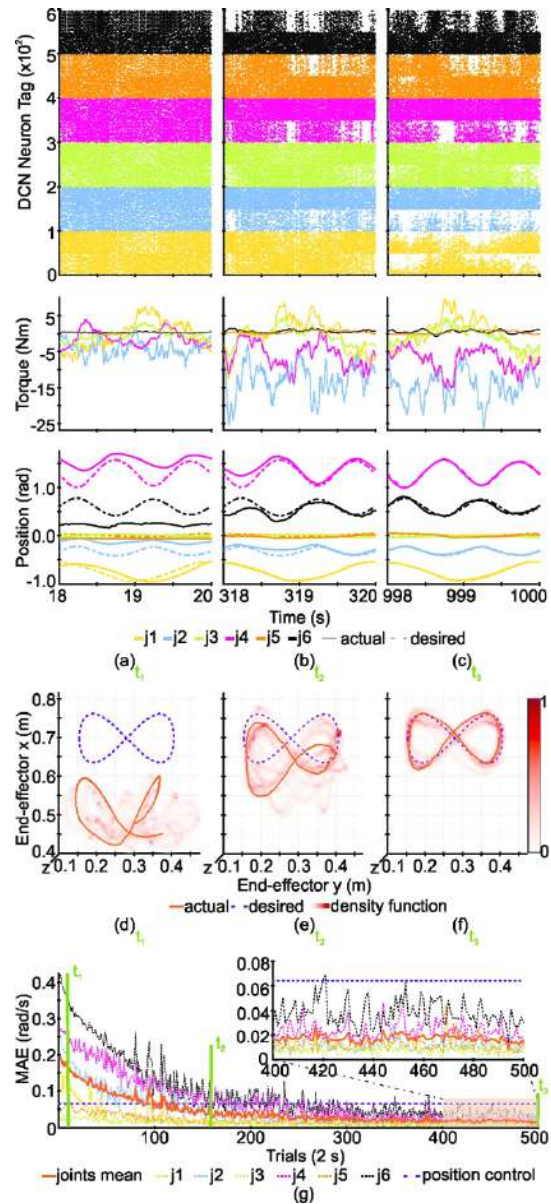

Fig. 5. Behavioural evolution through eight-like trajectory trials (2 s). (a) Initial learning stage (t1=18-20 s). (b) Intermediate learning stage (t2=318-320 s). (c) Final learning stage (t3=998-1000 s). The first row depicts the cerebellar output activity (DCN layer), whereas the second row shows its analogue conversion into torque commands. The third row illustrates the desired vs. actual trajectory per joint. (d), (e), and (f) reveal the desired vs. actual trajectory of the end-effector in Cartesian space at t1, t2, and t3 respectively. Also the density functions corresponding to the prior 10 trials are depicted. (g) Represents the position Mean Absolute Error (MAE) per trial through the learning process. The MAE of each joint is illustrated as well as the average MAE of all joints, completed with the default factory-installed position control baseline performance.

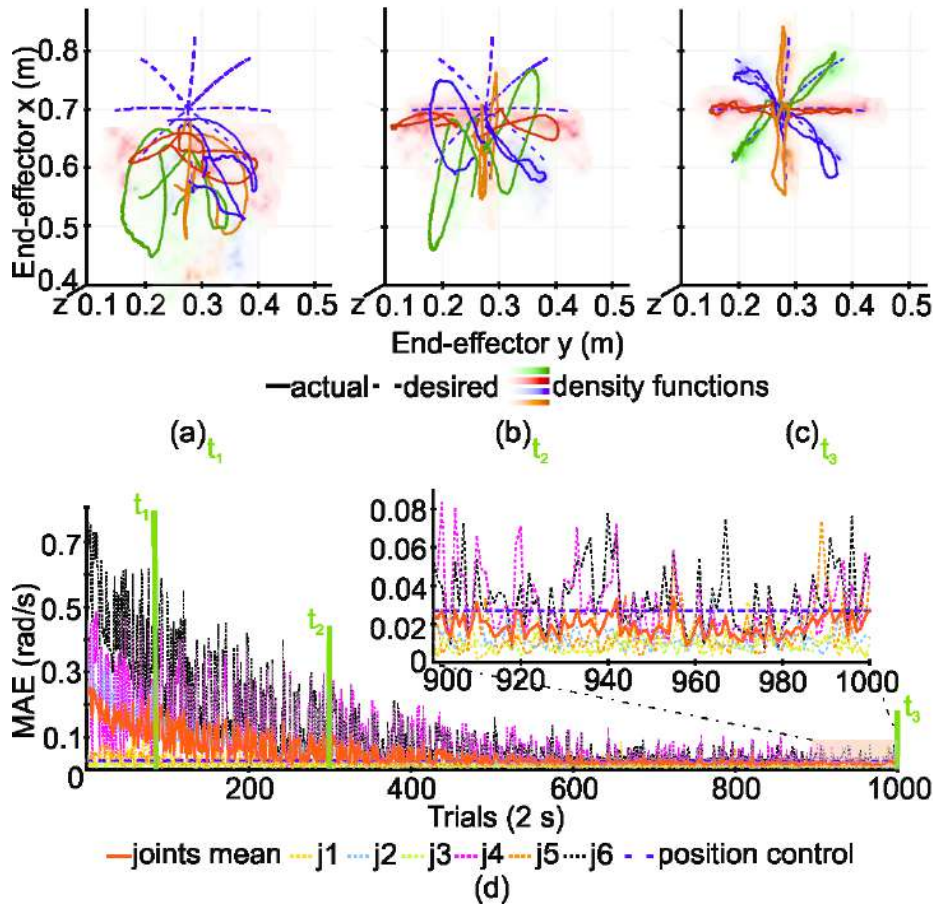

Fig. 6. Behavioural evolution through target reaching trials (2 s). Each trial consisted of one of the eight possible tasks. (a) Initial learning stage (t1=158-160 s). (b) Intermediate learning stage (t2=598-600 s). (c) Final learning stage (t3=1998-2000 s). (a), (b), and (c) depict the last performed trajectory for each of the eight possibilities in Cartesian space prior to t1, t2, and t3 respectively. The density functions reveal the end-effector behaviour over the last 80 trials, grouping the eight possible tasks by trajectory direction. (d) Represents the position Mean Absolute Error (MAE) per trial through the learning process. The MAE of each joint is illustrated as well as the mean MAE of all joints. High standard deviation values reflect how some reaching movements were more demanding than others. The position control baseline is the average MAE of the default factory-installed under the same stochastic distribution over trials.

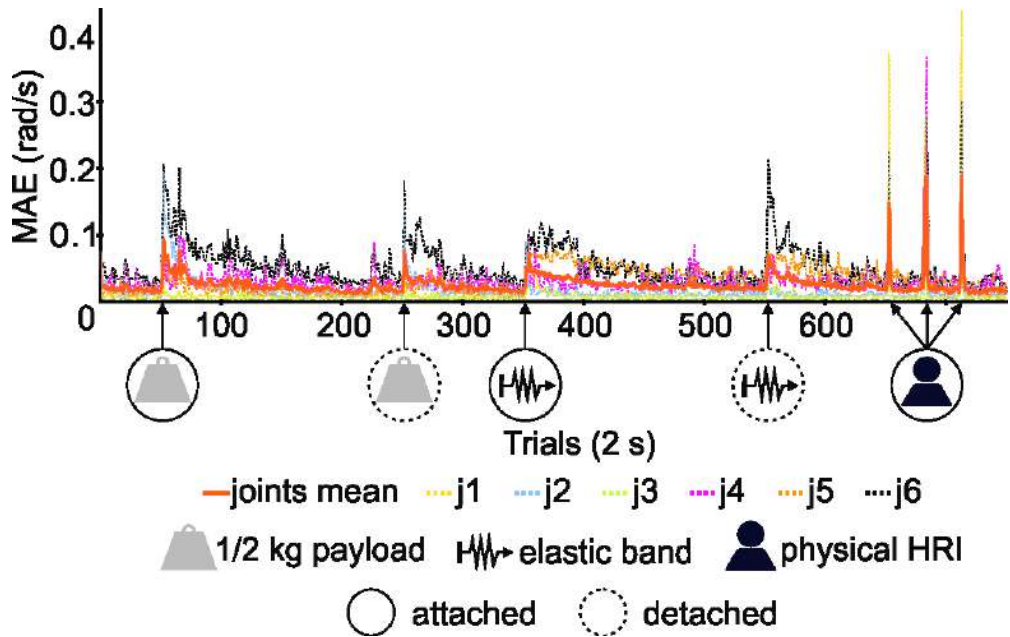

Fig. 7. Performance in an unstructured environment. Whilst performing the already learnt circular trajectory a set of unstructured interactions were undertaken: i) A ½ kg payload was attached to the end-effector and later on detached. ii) An elastic band was attached to the end-effector and later on detached. iii) A series of physical Human-Robot interactions. The figure depicts the position MAE through trials as interactions are undertaken, illustrating the cerebellar adaptation to unknown scenarios.